\documentclass{egpubl}
\usepackage{arxiv}

\ConferencePaper

\NoLicense

\usepackage[T1]{fontenc}
\usepackage{dfadobe}  
\usepackage{xspace} 
\usepackage{cite}

\BibtexOrBiblatex

\electronicVersion
\PrintedOrElectronic

\ifpdf \usepackage[pdftex]{graphicx} \pdfcompresslevel=9
\else \usepackage[dvips]{graphicx} \fi

\usepackage{egweblnk} 
\usepackage{multirow}
\usepackage{booktabs}
\usepackage{array}
\usepackage{amsmath,amssymb}

\newcommand{\methodname}{GeoFusionLRM\xspace}

\title[\methodname: Geometry-Aware Self-Correction for Consistent 3D Reconstruction]
      {\methodname: Geometry-Aware Self-Correction for Consistent 3D Reconstruction}

\author[A. Yildirim et al.]
{\parbox{\textwidth}{\centering
        Ahmet Burak Yildirim$^{1,*}$\orcid{0000-0003-3312-4280} 
        Tuna Saygin$^{1,*}$\orcid{0009-0009-8780-3028} 
        Duygu Ceylan$^{2}$\orcid{0000-0002-2307-9052} 
        Aysegul Dundar$^{1}$\orcid{0000-0003-2014-6325}
        }
        \\
{\parbox{\textwidth}{\centering 
         $^1$Bilkent University, Ankara, Turkey\\
         $^2$Adobe Research, London, United Kingdom\\[4pt]
         $^*$These authors contributed equally.
       }
}
}

\begin{document}

\maketitle

\begin{abstract}
Single-image 3D reconstruction with large reconstruction models (LRMs) has advanced rapidly, yet reconstructions often exhibit geometric inconsistencies and misaligned details that limit fidelity. We introduce \methodname, a geometry-aware self-correction framework that leverages the model’s own normal and depth predictions to refine structural accuracy. Unlike prior approaches that rely solely on features extracted from the input image, \methodname feeds back geometric cues through a dedicated transformer and fusion module, enabling the model to correct errors and enforce consistency with the conditioning image. This design improves the alignment between the reconstructed mesh and the input views without additional supervision or external signals. Extensive experiments demonstrate that \methodname achieves sharper geometry, more consistent normals, and higher fidelity than state-of-the-art LRM baselines. Project page: \url{https://geofusionlrm.abyildirim.com/}


\end{abstract}

\section{Introduction}

\begin{figure*}[!t]
\centering
\includegraphics[width=1.0\textwidth]{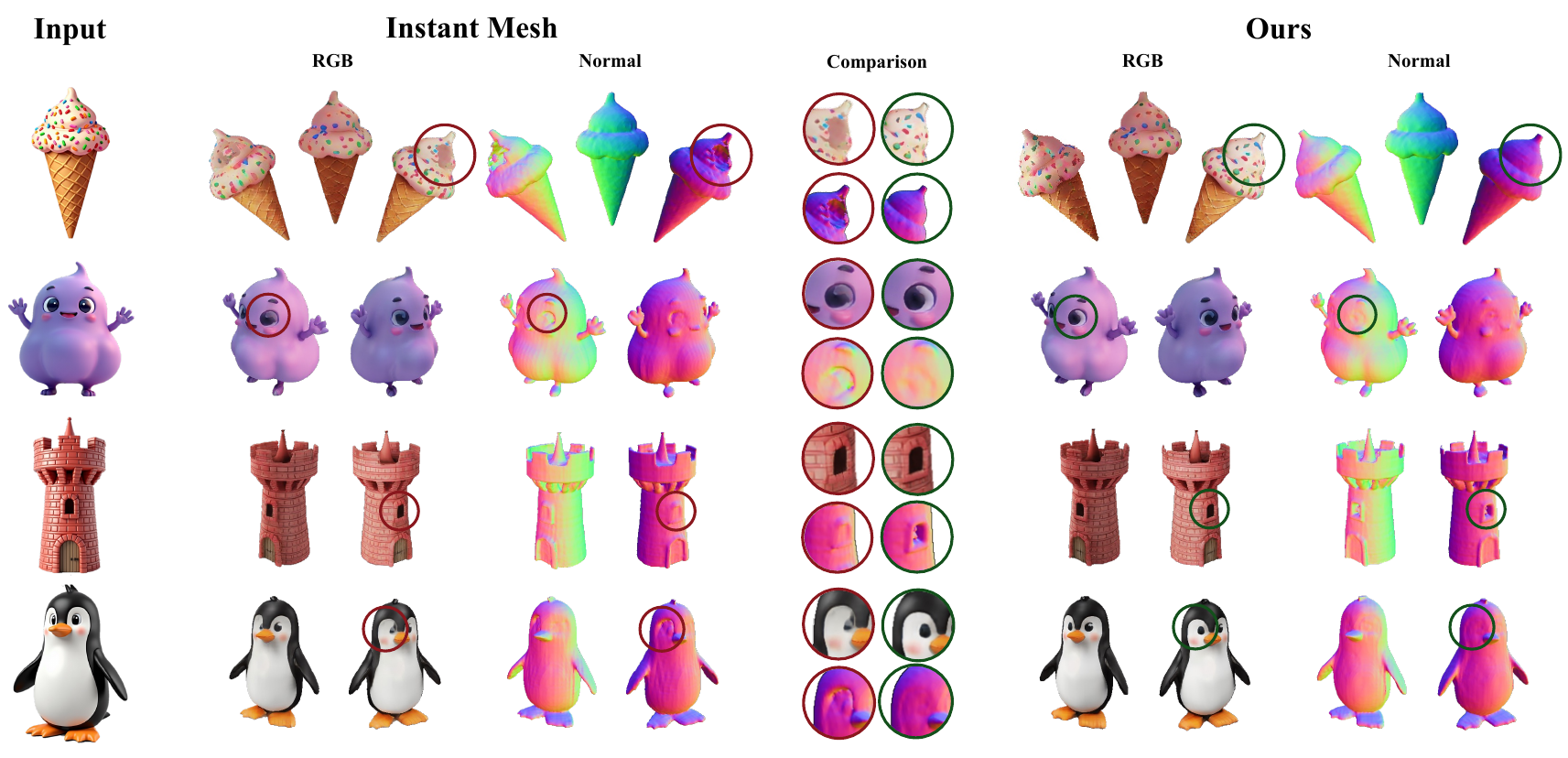}
\caption{Qualitative comparison using a synthesized input image generated by the FLUX image generator. The same synthesized image is provided as input to the InstantMesh baseline and our proposed \methodname. The baseline struggles to preserve geometric fidelity, producing distorted normals and misaligned surface details. In contrast, our iterative geometric conditioning progressively corrects these errors, yielding reconstructions with sharper normals and RGB renderings that more closely match the GT view.}
\label{fig:teaser}
\vspace{-10pt}
\end{figure*}

Recovering 3D geometry from images is fundamental to many applications in vision and graphics, such as content creation, AR/VR, and robotics. Reconstructing a full 3D model from a single image is particularly challenging due to the severe ambiguity of missing viewpoints. Recent Large Reconstruction Models (LRMs) \cite{hong2023lrm, wei2024meshlrm, li2023instant3d, wang2023pf, jiang2024real3d, xie2024lrm}, have made progress on this task by training transformer architectures on large collections of image–3D data pairs, enabling them to directly predict 3D assets from a single view. While these models succeed in generating 3D assets that capture the coarse 3D shape observed in the images, they often struggle to produce meshes that are consistent with the conditioning image in terms of geometric details. They suffer from limitations including inaccurate geometry, distorted surface normals, and misaligned details. These limitations highlight the need for approaches that improve consistency between the generated 3D assets and the input images. 

Existing approaches \cite{hong2023lrm, xu2024instantmesh, huang2025spar3d, tang2024lgm} typically encode 2D image features and inject them into the reconstruction pipeline via attention-level conditioning. While these models provide a strong semantic prior, accurate 3D reconstruction from a single image remains fundamentally ill-posed due to missing volumetric information. In particular, single-image methods such as LRM \cite{hong2023lrm} and SPAR3D \cite{huang2025spar3d} often struggle with occluded and back-facing regions, where geometry must be inferred from learned shape priors rather than direct visual evidence. On the other hand, recent LRM-based methods such as LGM \cite{tang2024lgm} and InstantMesh \cite{xu2024instantmesh} attempt to alleviate this limitation by leveraging pretrained image-to-multiview models \cite{shi2023mvdream, wang2023imagedream, liu2023zero} to synthesize multiple views from a single input image, which are then jointly encoded to provide a stronger geometric prior. However, the synthesized multi-view images are not always geometrically consistent. Moreover, when operating directly on real multi-view images without view synthesis, the available views are sometimes insufficiently informative when the object geometry is difficult to infer from RGB observations alone. As a result, ambiguities in 3D understanding can persist, which occasionally leads to reconstruction failures where geometric errors in the mesh are visually masked in the rendered RGB outputs (e.g., holes suggested by appearance but absent in the underlying geometry). To address this, our model predicts its own depth and normal maps from an initial reconstruction. These predicted geometric cues are then fused with image features in a second pass, allowing the network to refine depth, normals, and semantic information. This self-contained two-stage process improves geometric fidelity and enables more accurate and consistent mesh reconstruction without relying on external predictors. The improvements achieved by our approach on FLUX \cite{flux2024}-generated synthesized images are illustrated in Fig. \ref{fig:teaser}, demonstrating sharper geometry, more accurate normals, and better alignment with the input image compared to its baseline.

Building on this idea, we propose \methodname, a geometry-aware self-corrective conditioning framework that integrates these predicted geometric cues for two-stage refinement. Once an initial mesh is first reconstructed from the input image, we introduce a geometry encoder that encodes features of the depth and normal maps of this initial reconstruction. The resulting features are fused with semantic features from the vision encoder through a fusion module. This two-pass process corrects residual geometric errors and produces more accurate and consistent meshes. Our method builds upon InstantMesh \cite{xu2024instantmesh} as a baseline and introduces two key components: (i) a GeoFormer encoder, fine-tuned with geometric supervision to capture structural consistency from normals and depths, and (ii) the GeoFuser module, a lightweight token-wise network that fuses semantic features from the vision encoder with geometry-aware embeddings from GeoFormer. 

Our contributions can be summarized as follows:
\begin{itemize}
    \item \textbf{Self-predicted geometry-aware conditioning:} We introduce \methodname, which refines meshes in a two-stage process by conditioning on depth and normal cues predicted from intermediate reconstructions.
    \item \textbf{GeoFormer encoder:} We propose a geometry-aware encoder, initialized from DINO \cite{caron2021emerging} and fine-tuned with geometric supervision, to capture structural alignment with conditioning images.
    \item \textbf{GeoFuser module:} We design a lightweight token-level fusion network that merges semantic and geometric features to produce refined triplane conditioning.
    \item \textbf{Improved consistency and fidelity:} Extensive experiments show that \methodname improves over InstantMesh \cite{xu2024instantmesh} and other competing models, yielding sharper geometry, more accurate normals, and higher fidelity to input views.
\end{itemize}
\section{Related Work}

Single-image 3D reconstruction has advanced considerably with the advent of transformer-based architectures and large-scale training datasets. Earlier approaches predominantly employed category-specific encoder–decoder networks \cite{bhattad2021view, chen2019learning, goel2020shape, dundar2023fine, dundar2023progressive}, which restricted their ability to generalize beyond the categories seen during training. More recently, large reconstruction models (LRMs) have been introduced, trained on diverse collections of 3D assets \cite{hong2023lrm}, and shown to generate high-fidelity 3D geometry from sparse inputs such as a single image. These models shift the paradigm from category-specific learning toward general-purpose reconstruction, providing stronger robustness to variations in object shape and appearance. Architectures such as Instant3D \cite{li2023instant3d} and InstantMesh \cite{xu2024instantmesh} exemplify this trend by integrating multi-view diffusion \cite{shi2023zero123++} with transformer-based 3D decoders, producing triplane or volumetric representations that generalize across categories without requiring further fine-tuning. Recent extensions further improve efficiency and applicability, for example, through architectural simplifications and curated training data \cite{tochilkin2024triposr}, or by jointly estimating camera pose and geometry to enable reconstruction from unposed sparse inputs \cite{wang2023pf}. Collectively, these works highlight a transition from narrow, limited domain reconstructions toward more scalable and generalizable frameworks that form the basis for current research directions. However, despite their efficiency, existing LRMs remain sensitive to errors introduced during multi-view synthesis and lack explicit mechanisms for geometry-aware correction, causing geometric inconsistencies and misalignment with the conditioning image to persist once reconstruction is completed.

Several feed-forward approaches \cite{liu2023one, zheng2024mvd, li2024era3d, voleti2024sv3d} synthesize multi-view observations using diffusion-based view generation conditioned on a single image \cite{liu2023zero}, followed by 3D reconstruction through task-specific geometric pipelines rather than a unified LRM-style representation. As an extension, methods such as Wonder3D and GeoWizard \cite{long2024wonder3d, fu2024geowizard} jointly generate RGB images and surface normals, employing cross-domain attention mechanisms to ensure geometric consistency during view synthesis. Although these methods improve multi-view consistency, reconstruction quality remains constrained by inaccuracies in the synthesized views. In contrast, optimization-based approaches leverage strong 2D diffusion priors through per-scene procedures such as Score Distillation Sampling (SDS) \cite{tang2023make}, achieving high geometric fidelity at the cost of substantial computational overhead.

There have also been iterative approaches proposed for improving 3D reconstruction quality. Some methods refine reconstructions by progressively incorporating additional input views to update the underlying representation \cite{li2025lirm, kang2025ilrm}, which inherently assumes access to multi-view image sets. In contrast, our approach performs reconstruction from a single input image and applies refinement internally through predicted geometric cues rather than external view accumulation. Closely related, GTR \cite{zhuang2024gtr} introduces a lightweight per-instance refinement stage on top of Large Reconstruction Models, where NeRF color parameters and triplane features are fine-tuned at test time using differentiable mesh rendering. Its test-time optimization primarily targets appearance by refining RGB textures through color-based losses, without explicitly correcting the underlying geometry, in contrast to our geometry-aware refinement.

In another line of recent work, surface normals have been particularly used to improve geometric fidelity of reconstruction models \cite{patel2025normal, shen2025high}. These methods rely on an additional monocular depth or surface normal estimation network and provide normal maps as input to a reconstruction model directly. Our work differs in that we propose a two-stage self-corrective conditioning framework, which does not require additional networks or lengthy optimization steps.

\section{Method}

\begin{figure*}[!t]
\centering
\includegraphics[width=1.0\textwidth]{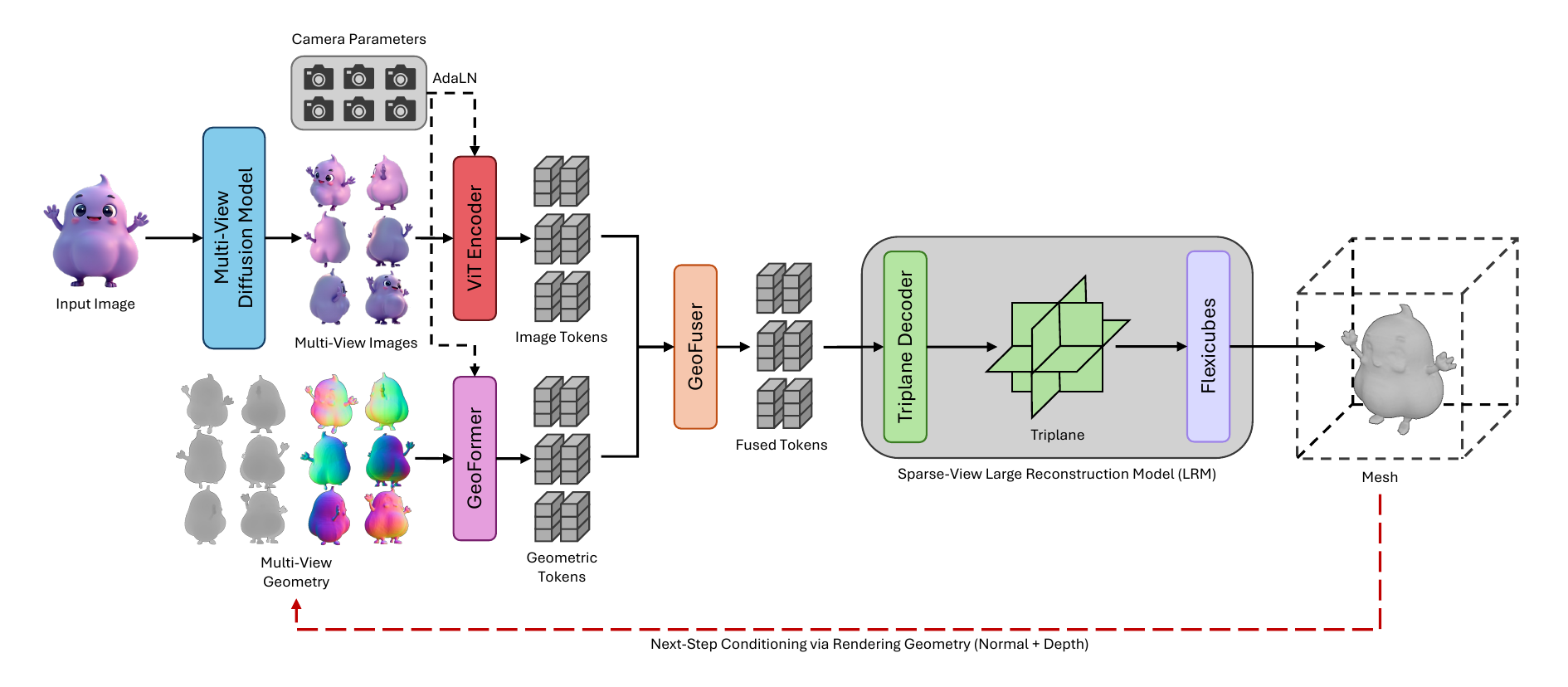}
\caption{Overview of the proposed \methodname architecture. Given a conditioning image, semantic features are extracted with a pre-trained vision encoder, while geometric cues from normals and depths of the intermediate mesh are encoded by the geometry-aware \textbf{GeoFormer}. The \textbf{GeoFuser} module merges these two streams of embeddings at the token level to produce refined conditioning features, which guide the LRM in generating an updated 3D mesh. This process corrects residual geometric errors and improves the consistency of surface normals and RGB renderings with respect to the conditioning image.}
\label{fig:architecture}
\vspace{-10pt}
\end{figure*}

Our goal is to improve the geometric consistency of single-image 3D reconstruction with respect to the conditioning image. To this end, we introduce \textbf{\methodname}, an iterative conditioning framework that augments Large Reconstruction Models (LRMs) with geometry-aware self-supervision. The overall architecture is illustrated in Fig. \ref{fig:architecture}. We first review the baseline LRM pipeline before describing our geometry-aware extensions.

\subsection{Preliminaries}

Large Reconstruction Models (LRMs) such as InstantMesh \cite{xu2024instantmesh} predict a 3D representation directly from a single conditioning image by employing a transformer-based architecture with 3D-aware cross-attention. In practice, InstantMesh first employs a multi-view diffusion model \cite{shi2023zero123++} to synthesize six views of the object $\{I_k\}_{k=1}^{6}$ from the input image $I$, each associated with known camera parameters $\{C_k\}_{k=1}^{6}$. These views are then encoded by a transformer backbone (ViT Encoder) with DINO \cite{caron2021emerging} initialization, where the camera parameters are injected into the AdaLN layers. The encoded semantic tokens are defined as
\begin{equation}
    F^{\text{sem}}_k = E_{\text{sem}}(I_k, C_k),
    \label{eq:instantmesh_sem}
\end{equation}
where $F^{\text{sem}}_k \in \mathbb{R}^{N \times d}$ denotes the semantic token embeddings for view $k$. The triplane decoder transformer aggregates these tokens across views and generates the triplane tokens by applying cross-attention between the semantic tokens and triplane tokens. The resulting features form a triplane representation $\mathcal{T} \in \mathbb{R}^{3 \times R \times R \times d}$ are subsequently decoded into a mesh via differentiable iso-surface extraction. Although this design provides stronger 3D awareness than earlier LRMs, the reliance on semantic embeddings still limits geometric fidelity. While camera parameters can inject 3D information into the 2D tokens, they often fail to capture structural details, leading to unwanted artifacts in the reconstructed geometry. This embedding bottleneck motivates our geometric refinement strategy, where the model receives feedback from geometry rendered under the input view to progressively refine the embeddings extracted from RGB images.

\subsection{\methodname Overview}

\methodname introduces geometry-aware conditioning into the LRM pipeline (see Fig. \ref{fig:architecture}). Starting from an initial mesh $\mathcal{M}^{(0)}$ generated by the baseline LRM, we extract depth and normal maps:
\begin{equation}
    D^{(t)}, \; N^{(t)} = \Pi(\mathcal{M}^{(t)}),
    \label{eq:render_normals_depth}
\end{equation}
where $\Pi(\cdot)$ denotes differentiable rendering of the current mesh $\mathcal{M}^{(t)}$ into depth $D$ and normals $N$ as defined in Eq. \ref{eq:render_normals_depth}. These maps are encoded by the geometry-aware \textbf{GeoFormer}, and fused with semantic tokens through the \textbf{GeoFuser} module to guide the next reconstruction pass.

\subsection{GeoFormer: Geometry-Aware Encoder}
GeoFormer is designed to capture structural consistency from normal and depth projections. It is initialized as a copy of the ViT encoder used in InstantMesh, including AdaLN-based camera parameter conditioning. To process normal and depth information, we extend the input layer from three channels (RGB) to four channels with zero-initialized weights, which enables the encoder to handle geometry maps instead of the color input. Formally, the geometry-aware tokens are defined as
\begin{equation}
    F^{\text{geo}} = E_{\text{geo}}(D^{(t)}, N^{(t)}),
    \label{eq:geoformer_tokens}
\end{equation}
where $E_{\text{geo}}$ denotes the GeoFormer encoder and $F^{\text{geo}} \in \mathbb{R}^{N \times d}$ are geometry-aware embeddings. GeoFormer is added as a parallel branch to the pipeline, which produces geometry-aware tokens that are later fused with semantic tokens from the vision encoder through the GeoFuser module.

\subsection{GeoFuser: Token-Level Feature Fusion}
To integrate semantic and geometric features, we propose the \textbf{GeoFuser} module. Given semantic tokens $F^{\text{sem}}$ from the input image and geometry-aware tokens $F^{\text{geo}}$ extracted from rendered normal and depth maps of the same view (Eq.~\ref{eq:render_normals_depth} and Eq.~\ref{eq:geoformer_tokens}), GeoFuser produces corrective residuals that refine the semantic embeddings. Formally, the fused tokens are defined as
\begin{equation}
    F^{\text{fused}} = F^{\text{sem}} + f_\theta(F^{\text{sem}}, F^{\text{geo}}),
    \label{eq:geofusion_summary}
\end{equation}
where $f_\theta(\cdot)$ is a lightweight two-layer feed-forward network with a hidden SiLU activation. Its final linear layer is initialized with zero weights and bias, ensuring that the residual correction is disabled at the first iteration and enabled in the subsequent refinement. At the first iteration, the residual term is disabled, and the baseline LRM generates an initial mesh. In the following iteration, geometry maps rendered from the reconstructed mesh are encoded by GeoFormer and fused through Eq.~\ref{eq:geofusion_summary} to refine the semantic embeddings based on inconsistencies between input image features and rendered geometry. The refined tokens $F^{\text{fused}}$ are then injected into the cross-attention layers of the LRM, subsequently improving geometric fidelity.

During training, we unroll the refinement process for $T = 3$ steps. At each step $t \in \{0, 1, 2\}$, the model renders the current reconstruction, encodes the resulting depth and normal maps, and receives supervision on the outputs for that step. This setup makes the optimization stable and keeps the overall training cost manageable. At inference time, however, we observe that an initial reconstruction stage followed by a single refinement stage is sufficient.

\subsection{Training Strategy}
During training, the baseline InstantMesh backbone is frozen, and only the parameters of \emph{GeoFormer} and \emph{GeoFuser} are optimized.

For supervision, we utilize the default InstantMesh losses on the rendered outputs and backpropagate immediately. In particular, this objective combines photometric MSE, perceptual LPIPS \cite{zhang2018perceptual}, mask consistency, depth alignment, normal similarity, and FlexiCubes \cite{shen2023flexicubes} regularization, with the same weights as InstantMesh:
\begin{equation}
\begin{split}
    \mathcal{L} = \;& \mathcal{L}_{\text{rgb}}
    + 2.0 \cdot \mathcal{L}_{\text{lpips}}
    + \mathcal{L}_{\text{mask}} \\
    &+ 0.5 \cdot \mathcal{L}_{\text{depth}}
    + 0.2 \cdot \mathcal{L}_{\text{normal}}
    + \mathcal{L}_{\text{reg}}.
\end{split}
\label{eq:training_loss}
\end{equation}
\section{Experiments}

In this section, we present quantitative and qualitative results for \methodname, detailing our training setup, evaluation datasets, baselines, and metrics.

\noindent\textbf{Datasets.}\label{sec:exp-datasets} We train our model on Objaverse-1.0 \cite{deitke2023objaverse}, excluding assets with rendered alpha coverage of $\leq 10\%$ to remove highly transparent or very small objects. After filtering, the training set contains approximately $168k$ objects. For evaluation, we use the OmniObject3D \cite{wu2023omniobject3d} and Google Scanned Objects (GSO) \cite{downs2022google} datasets. From OmniObject3D, we select five objects per category across $100$ categories ($500$ in total), while we uniformly sample $500$ objects from GSO. Each object is rendered on a fixed viewing grid, defined by elevations $\{-20^\circ,-10^\circ,0^\circ,10^\circ,20^\circ\}$ crossed with six uniformly spaced azimuths in an orbital setting. In addition to these predefined views, we evaluate our method on the standard OmniObject3D benchmark test views, following the evaluation protocol used in prior work (e.g., InstantMesh). Specifically, we report results on $16$ benchmark views per object provided by the OmniObject3D dataset.

\noindent\textbf{Implementation Details.}
We fine-tune our model for $168k$ steps on $4~\times$ A100 GPUs using the AdamW optimizer with an initial learning rate of $4\times10^{-6}$, $\beta_1=0.90$, $\beta_2=0.95$, and a weight decay of $0.01$. We employ a cosine scheduler that anneals the learning rate to $0$ over $100k$ steps. Training supervision is provided with $32$ views per object, obtained by randomly sampling camera poses on a viewing sphere with radius $r \sim \mathcal{U}(1.5,\,2.2)$ and uniformly distributed orientations. 

\noindent\textbf{Baselines.} 
We compare \methodname against four recent approaches: LRM \cite{hong2023lrm}, SPAR3D \cite{huang2025spar3d}, LGM \cite{tang2024lgm}, and InstantMesh \cite{xu2024instantmesh}. OpenLRM \cite{openlrm} weights are utilized for LRM comparison, which is an open-source implementation of the Large Reconstruction Model (LRM) \cite{hong2023lrm}. It is a transformer-based framework trained to predict NeRF representations of the object from a single input image. SPAR3D follows a two-stage design: it first predicts a sparse 3D point cloud with a lightweight point-diffusion model, and then refines it into a detailed mesh conditioned on the input view. LGM reconstructs 3D objects as collections of Gaussian parameters for an efficient representation that can predict high-quality novel view synthesis. InstantMesh integrates an off-the-shelf multi-view diffusion model with an LRM-style sparse-view reconstructor, enabling fast feed-forward mesh generation from a single image.

\noindent\textbf{Metrics.}
We evaluate both the visual quality and geometric accuracy of the generated 3D assets. For visual quality, we compare predicted and ground-truth pixel values in RGB space, reporting PSNR, SSIM, and LPIPS (higher is better for PSNR/SSIM, while lower is better for LPIPS). For geometry, we assess structural consistency by rendering normal maps from the generated meshes in Blender using the same camera grid, and then comparing predicted and ground-truth normals with the same set of metrics. All results are averaged over both views and objects.

\begin{figure*}[!t]
\centering
\includegraphics[width=1.0\textwidth]{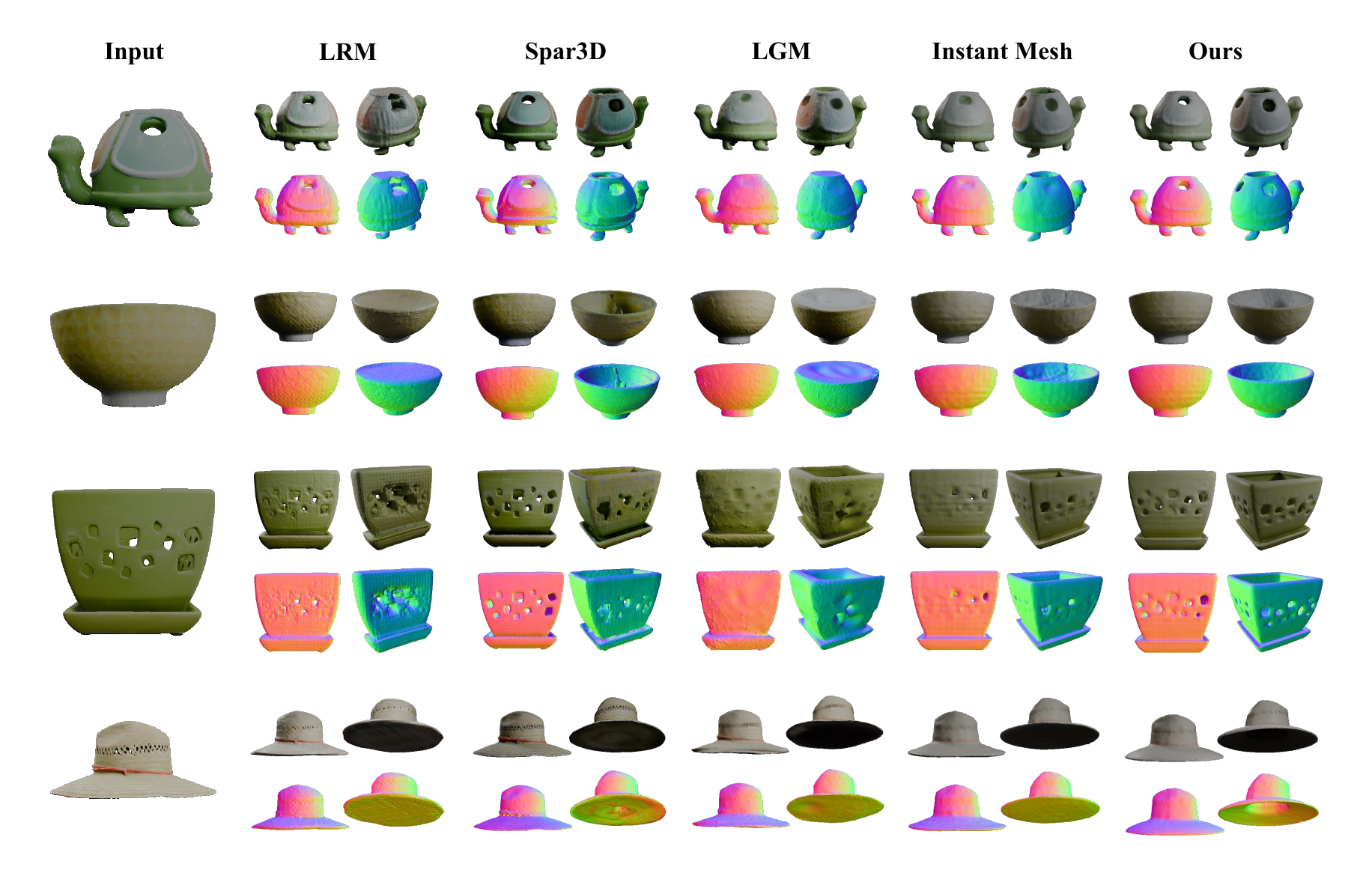}
\caption{Qualitative results on GSO.
Columns show the conditioning input image (left), followed by LRM, SPAR3D, LGM, InstantMesh, and our \methodname. For each method, we display results rendered from the same camera viewpoints, showing RGB outputs (top) and surface normals (bottom).}
\label{fig:sota_comp}
\vspace{-10pt}
\end{figure*}

\begin{table}[h]
\centering
\caption{Quantitative results on the \textbf{GSO dataset using uniform views} for RGB images and normal maps.}
\label{tab:quant_gso}
\setlength{\tabcolsep}{6pt}
\scriptsize
\begin{tabular}{l ccc ccc}
\toprule
\multirow{2}{*}{Method} & \multicolumn{3}{c}{RGB} & \multicolumn{3}{c}{Normal} \\
\cmidrule(lr){2-4}\cmidrule(lr){5-7}
& PSNR $\uparrow$ & SSIM $\uparrow$ & LPIPS$\downarrow$ & PSNR$\uparrow$ & SSIM$\uparrow$ & LPIPS $\downarrow$\\
\midrule
OpenLRM      & 17.22 & 0.889 & 0.1212 & 18.74 & 0.884 & 0.1243 \\
Spar3d       & 16.15 & 0.879 & 0.1214 & 19.05 & 0.900 & 0.1160 \\
LGM          & 19.09 & 0.902 & 0.0975 & 22.08 & 0.921 & 0.0886 \\
InstantMesh  & 20.31 & 0.920 & 0.0832 & 25.83 & 0.947 & 0.0625 \\
Ours & \textbf{20.35} & \textbf{0.921} & \textbf{0.0831} & \textbf{26.39} & \textbf{0.950} & \textbf{0.0592} \\
\bottomrule
\end{tabular}
\vspace{-10pt}
\end{table}

\begin{table}[h]
\centering
\caption{
Quantitative results on \textbf{OmniObject3D using uniform views} for RGB images and normal maps.}
\label{tab:quant_o3d}
\setlength{\tabcolsep}{6pt}
\scriptsize
\begin{tabular}{l ccc ccc}
\toprule
\multirow{2}{*}{Method} & \multicolumn{3}{c}{RGB} & \multicolumn{3}{c}{Normal} \\
\cmidrule(lr){2-4}\cmidrule(lr){5-7}
& PSNR $\uparrow$ & SSIM $\uparrow$ & LPIPS$\downarrow$ & PSNR$\uparrow$ & SSIM$\uparrow$ & LPIPS $\downarrow$\\
\midrule
OpenLRM      & 16.60 & 0.875 & 0.1189 & 19.30 & 0.870 & 0.1191 \\
Spar3d       & 17.08 & 0.885 & 0.1048 & 19.65 & 0.894 & 0.1092 \\
LGM          & 20.60 & 0.906  & 0.0819 & 23.31 & 0.905 & 0.0831 \\
InstantMesh  & 21.94 & 0.915 & 0.0798 & 24.68 & 0.918 & 0.0781 \\
Ours   & \textbf{23.05} & \textbf{0.921} & \textbf{0.0722} & \textbf{26.16} & \textbf{0.927} & \textbf{0.0658} \\
\bottomrule
\end{tabular}
\vspace{-10pt}
\end{table}

\begin{table}[h]
\centering
\caption{Quantitative results on \textbf{OmniObject3D using benchmark views} for RGB images and normal maps.}
\label{tab:quant_o3d_benchmark}
\setlength{\tabcolsep}{6pt}
\scriptsize
\begin{tabular}{l ccc ccc}
\toprule
\multirow{2}{*}{Method} & \multicolumn{3}{c}{RGB} & \multicolumn{3}{c}{Normal} \\
\cmidrule(lr){2-4}\cmidrule(lr){5-7}
& PSNR $\uparrow$ & SSIM $\uparrow$ & LPIPS $\downarrow$
& PSNR $\uparrow$ & SSIM $\uparrow$ & LPIPS $\downarrow$ \\
\midrule
SPAR3D        & 16.15 & 0.881 & 0.1235 & 17.52 & 0.883 & 0.1248 \\
LRM           & 16.85 & 0.874 & 0.1261 & 18.04 & 0.865 & 0.1245 \\
LGM           & 20.98 & 0.905 & 0.0849 & 22.61 & 0.906 & 0.0826 \\
InstantMesh   & 21.85 & 0.913 & 0.0805 & 24.24 & 0.918 & 0.0769 \\
Ours          & \textbf{22.75} & \textbf{0.916} & \textbf{0.0741}
              & \textbf{25.76} & \textbf{0.926} & \textbf{0.0648} \\
\bottomrule
\end{tabular}
\vspace{-10pt}
\end{table}

\subsection{Quantitative Comparisons}

Tables \ref{tab:quant_gso}, \ref{tab:quant_o3d}, and \ref{tab:quant_o3d_benchmark} summarize quantitative results on the GSO and OmniObject3D datasets. Across both datasets, \methodname consistently ranks first in all metrics, especially in the normal map scores.

In GSO, improvements are more pronounced in the normal metrics than in RGB, which reflects the geometry-focused conditioning of our pipeline. In OmniObject3D, the same pattern is observed for both uniform and benchmark views: RGB results show a modest improvement, whereas normal maps demonstrate consistently larger improvements across all metrics, which indicates improved geometric accuracy, especially for objects with thin structures and curved surfaces. Furthermore, we outperform the strongest baseline, InstantMesh, both qualitatively and quantitatively.

\subsection{Qualitative Comparisons}

Fig. \ref{fig:sota_comp} provides a qualitative comparison against state-of-the-art methods on the GSO dataset. These results reveal a critical weakness in several prior models: the tendency to prioritize plausible-looking textures at the expense of geometric fidelity. This is often achieved through aggressive surface smoothing and by baking complex shading cues directly into the albedo. While this can mask underlying shape defects in the RGB output, it leads to significantly distorted and uninformative surface normals.

This failure mode is particularly evident in the reconstructions from LGM and InstantMesh. For instance, with the turtle teapot, both methods flatten the complex pattern on the shell into a smooth, featureless surface. Similarly, for the bowl, the sharp rim is incorrectly rounded off. This geometric simplification is noticeably visible in their corresponding normal maps, which lack high-frequency detail. The most telling example is the decorative planter, where the crisp, cut-out patterns are degraded into shallow, indistinct indentations, demonstrating a failure to reconstruct complex topology.

In contrast, SPAR3D and our method, \methodname, show a superior capability to preserve fine geometric details. Both methods successfully capture the sharp edges of the bowl's rim and the complex holes of the planter, resulting in crisp, well-defined normal maps that accurately reflect the object's true surface structure.

However, our method also improves on reconstructing dark, low-feature regions. This is best illustrated by the hat reconstruction in the final row. The dark band on the hat presents an ambiguous region for reconstruction. While most methods produce a simplified shape, SPAR3D hallucinates a large hole, failing to infer the continuous surface underneath the dark texture. \methodname, on the other hand, correctly reconstructs the complete, coherent geometry of the hat, demonstrating a superior ability to reason about shape even in areas with ambiguous visual information. This highlights our model's improved capacity for producing not only detailed but also geometrically complete reconstructions.

\subsection{Ablation Studies}
\label{sec:ablation}

We conduct a comprehensive ablation analysis to evaluate the contribution of each design choice in our method. Specifically, we study the effects of input conditioning, architectural components, and the initialization strategy of the geometry-aware encoder. All ablations are evaluated using SSIM and LPIPS.

Table \ref{tab:ablation_combined} reports ablation results evaluated on both RGB and normal map reconstruction quality. Using only depth or only normal conditioning consistently degrades performance compared to using both conditions jointly, which confirms that depth and normal information provide complementary geometric signals. Replacing the proposed geometry-aware fusion with simple token concatenation or removing the GeoFormer initialization leads to a noticeable drop in reconstruction quality. While these variants still outperform the InstantMesh baseline, they remain below the proposed method across all metrics. The overall trends are consistent across RGB and normal evaluations, with the proposed method achieving the best performance in both RGB and normal reconstructions.

\begin{table}[t]
\centering
\caption{Ablation results evaluated on RGB and normal map reconstruction quality.}
\label{tab:ablation_combined}
\begin{tabular}{lcccc}
\toprule
 & \multicolumn{2}{c}{RGB} & \multicolumn{2}{c}{Normal} \\
\cmidrule(lr){2-3} \cmidrule(lr){4-5}
Method 
& SSIM $\uparrow$ & LPIPS $\downarrow$ & SSIM $\uparrow$ & LPIPS $\downarrow$ \\
\midrule
InstantMesh & 0.915 & 0.0798 & 0.918 & 0.0781 \\
Random Init & 0.914 & 0.0755 & 0.924 & 0.0680 \\
Token Concat & 0.916 & 0.0739 & 0.926 & 0.0663 \\
Normal Only & 0.916 & 0.0738 & 0.926 & 0.0662 \\
Depth Only & 0.916 & 0.0738 & 0.926 & 0.0661 \\
\textbf{Ours (Proposed)} & \textbf{0.920} & \textbf{0.0722} & \textbf{0.927} & \textbf{0.0658} \\
\bottomrule
\end{tabular}
\vspace{-10pt}
\end{table}

Finally, we analyze the effect of the initialization strategy. Initializing GeoFormer from random weights leads to consistent drops in SSIM and LPIPS in both RGB and normal evaluations. In contrast, initializing from a pretrained ViT encoder provides a stronger and more stable starting point for learning geometry-aware features. Since the DINO encoder used in InstantMesh is trained on a large dataset with geometry-aware adaptive normalization through camera parameters, reusing this encoder for the conditioning branch smooths the optimization landscape and allows the refinement stage to focus on correcting geometric inconsistencies rather than learning geometry-aware 2D-to-3D feature mappings from scratch.

\subsection{Iteration Analysis}

\begin{figure}[t]
  \centering
  \includegraphics[width=\columnwidth]{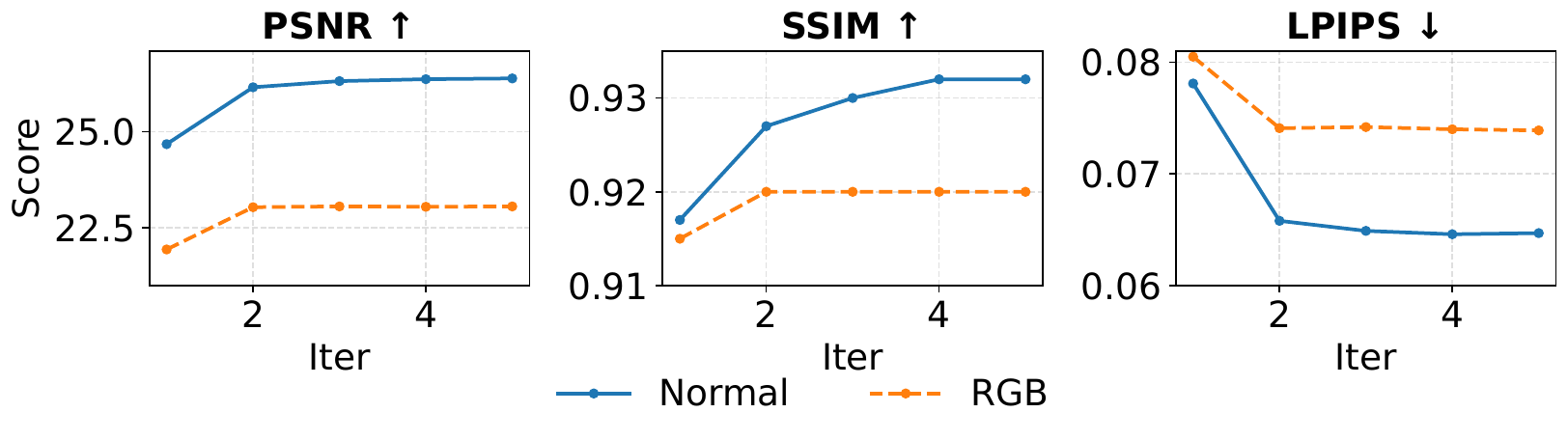}
  \caption{Performance across refinement iterations on the OmniObject3D dataset under uniform views.}
  \label{fig:iter_metrics}
  \vspace{-10pt}
\end{figure}

\methodname's ability to condition on its own rendered depth and normal maps enables iterative refinement of the generated geometry. To assess the effectiveness of this strategy, we evaluate performance across varying numbers of refinement steps, where Iteration 1 corresponds to a single forward pass without geometric conditioning. As shown in Fig.  \ref{fig:iter_metrics}, both geometric and appearance metrics (PSNR, SSIM, LPIPS) improve substantially after the first refinement pass, while additional iterations yield diminishing returns and quickly plateau. Considering the computational cost of each pass (Sec. \ref{subsec:cost}), we adopt a two-iteration setting, one initial reconstruction followed by one geometry-aware refinement, as the optimal trade-off between quality and efficiency.

\subsection{Computational Cost Analysis}

As refinement requires additional passes, we evaluate the inference-time computational cost of \methodname in comparison to InstantMesh. Table \ref{tab:computational_cost} reports TFLOPs and single-object inference time measured on an NVIDIA RTX 3090, where the same object is reconstructed by both methods. While our method improves reconstruction quality through geometry-aware refinement, this improvement comes with a higher computational cost. Specifically, the refinement process introduces extra forward passes over the base InstantMesh encoder, resulting in increased TFLOPs and longer inference time. This highlights the trade-off between reconstruction quality and computational efficiency for refinement-based methods.
\section{Conclusion}

We have presented \methodname, which aims to improve geometric fidelity of large reconstruction models through a refinement strategy. In particular, we introduce a geometric feature encoder and fusion modules into a baseline LRM method (InstantMesh in our experiments). Given an initial 3D reconstruction, these modules extract geometric features from the depth and normal maps of the reconstruction that act as residual features and improve the reconstruction. Our experiments show that this approach improves upon the base LRM method, especially in terms of geometric fidelity. The gains in geometric fidelity are most evident on surfaces exhibiting coherent shape deviations, such as flattened bumps, softened edges, and merged or missing holes, where geometric errors persist consistently across a region.

\begin{table}[t]
\caption{Computational cost and inference time comparison between InstantMesh and \methodname.}
\centering
\begin{tabular}{lcc}
\toprule
Model & TFLOPs $\downarrow$ & Inference Time $\downarrow$ \\
\midrule
InstantMesh & 3.878 & 0.989 \\
Ours & 8.687 & 3.854 \\
\bottomrule
\end{tabular}
\label{tab:computational_cost}\label{subsec:cost}
\end{table}

\begin{figure}[h]
  \centering
  \includegraphics[width=\columnwidth]{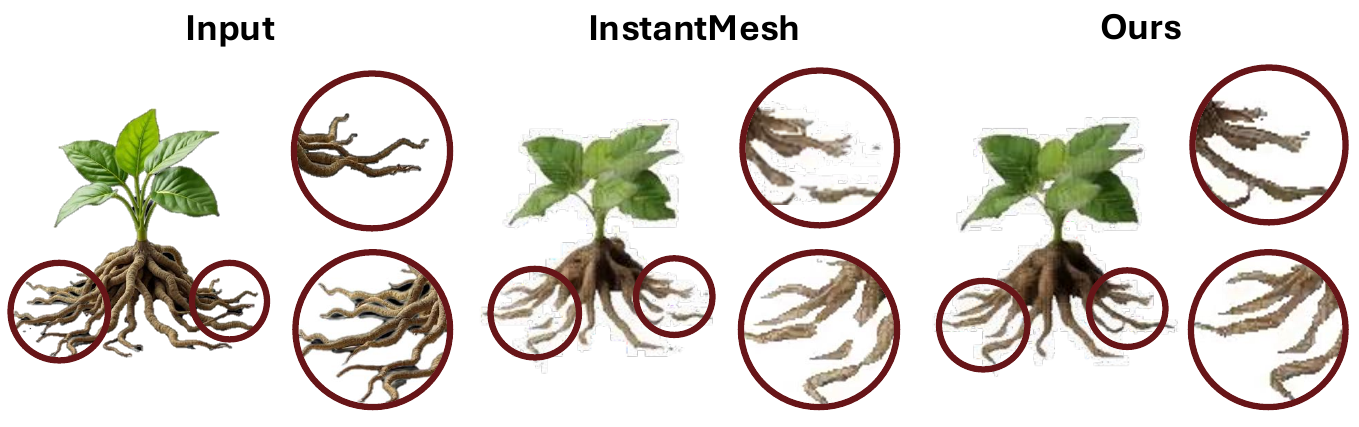}
  \caption{Limitations on thin structure reconstruction. Our refinement improves coarse branches by closing gaps (see zooms), but very thin root segments remain missing due to the limited resolution of the InstantMesh triplane backbone.}
  \label{fig:limitations}
  \vspace{-8pt}
\end{figure}

\textbf{Limitations.} As discussed in Sec. \ref{subsec:cost}, the proposed refinement strategy increases inference-time computation due to additional forward passes. In terms of reconstruction quality, thin structures such as small branches and fine root segments remain a challenging case for LRM-based models. Geometry-conditioned self-refinement improves branching structures for the plant object shown in Fig. \ref{fig:limitations} compared to InstantMesh. The refinement effectively fixes gaps and discontinuities observed in the InstantMesh reconstruction along coarse root branches, resulting in improved structural coherence and surface continuity. However, very fine root branches remain missing in both the baseline and our result. This limitation stems from the low-resolution triplane representation used by the InstantMesh backbone, which restricts the recovery of extremely fine-scale geometric details. Overall, while the refinement step consistently improves branch coherence and corrects large-scale structural errors, details below the effective resolution of the underlying representation remain challenging for both methods.

\textbf{Future work.} Future research could extend our approach beyond local geometric cues like depth and normals. One promising direction is to incorporate global geometric priors, including symmetry constraints or semantic structural consistency, directly into the fusion module to further regularize the reconstruction.

\section{Acknowledgments}
This work was supported by the BAGEP Award of the Science Academy.
We acknowledge the EuroHPC Joint Undertaking for awarding the project ID
EHPC-BEN-2025B05-045 access to the MareNostrum5 ACC at Barcelona, Spain.
The author also acknowledges support from the Scientific and Technological Research Council of Turkey (TUBITAK) through the 2211 Graduate Scholarship Program.

\bibliographystyle{eg-alpha-doi} 
\bibliography{egbibsample}

\end{document}